# Meta-Learning Evolutionary Artificial Neural Networks

Ajith Abraham
Department of Computer Science, Oklahoma State University
700 N Greenwood Avenue, Tulsa, OK 74106-0700, USA
Email: ajith.abraham@ieee.org, URL: http://ajith.softcomputing.net

**Abstract**

In this paper, we present MLEANN (Meta-Learning Evolutionary Artificial Neural Network), an automatic computational framework for the adaptive optimization of artificial neural networks wherein the neural network architecture, activation function, connection weights; learning algorithm and its parameters are adapted according to the problem. We explored the performance of MLEANN and conventionally designed artificial neural networks for function approximation problems. To evaluate the comparative performance, we used three different well-known chaotic time series. We also present the state of the art popular neural network learning algorithms and some experimentation results related to convergence speed and generalization performance. We explored the performance of backpropagation algorithm; conjugate gradient algorithm, quasi-Newton algorithm and Levenberg-Marquardt algorithm for the three chaotic time series. Performances of the different learning algorithms were evaluated when the activation functions and architecture were changed. We further present the theoretical background, algorithm, design strategy and further demonstrate how effective and inevitable is the proposed MLEANN framework to design a neural network, which is smaller, faster and with a better generalization performance.

**Key words:** global optimization, local search, evolutionary algorithm and meta-learning

## 1. Introduction

The strong interest in neural networks in the scientific community is fueled by the many successful and promising applications especially to tasks of optimization [26], speech recognition [16], pattern recognition [12], signal processing [57], function approximation [79], control problems [3] [5], financial modeling [67] etc.. Even though artificial neural networks are capable of performing a wide variety of tasks, yet in practice sometimes they deliver only marginal performance. Inappropriate topology selection and learning algorithm are frequently blamed. There is little reason to expect that one can find a uniformly best algorithm for selecting the weights in a feedforward artificial neural network. This is in accordance with the no free lunch theorem, which explains that for any algorithm, any elevated performance over one class of problems is exactly paid for in performance over another class [54]. In sum, one should be skeptical of claims in the literature on training algorithms that one being proposed is substantially better than most others. Such claims are often defended through some simulations based on applications in which the proposed algorithm performed better than some familiar alternative [41] [44] [45] [78].

At present, neural network design relies heavily on human experts who have sufficient knowledge about the different aspects of the network and the problem domain. As the complexity of the problem domain increases, manual design becomes more difficult and unmanageable. Evolutionary design of artificial neural networks has been widely explored. Evolutionary algorithms are used to adapt the connection weights, network architecture and learning rules according to the problem environment. A distinct feature of evolutionary neural networks is their adaptability to a dynamic environment. In other words, such neural networks can adapt to an environment as well as changes in the environment. The two forms of adaptation: evolution and learning in evolutionary artificial neural networks make their adaptation to a dynamic environment much more effective and efficient than the conventional learning approach [2]. In Section 2, we present the different neural network learning paradigms followed by some experimentation results to demonstrate the difficulties to design neural networks, which are smaller, faster and with a better generalization performance. In section 3, we introduce evolutionary algorithms and state of the art design of Evolutionary Artificial Neural Networks (EANNs) followed by the proposed MLEANN framework [2]. In the MLEANN framework, in addition to the evolutionary search of connection weights and architectures (connectivity and activation functions), local search techniques are used to fine-tune the weights (meta-learning). Experimentation results are provided in Section 3.3.1 and some discussions and conclusions are provided towards the end.

## 2. Artificial Neural Network Learning Algorithms

The artificial neural network (ANN) methodology enables us to design useful nonlinear systems accepting large numbers of inputs, with the design based solely on instances of input-output relationships. For a training set $T$ consisting of $n$ argument value pairs and given a $d$-dimensional argument $x$ and an associated target value $t$ will be approximated by the neural network output. The function approximation could be represented as

$T = \{(x_i, t_i) : i = 1 : n\}$

In most applications the training set *T* is considered to be noisy and our goal is not to reproduce it exactly but rather to construct a network function that generalizes well to new function values. We will try to address the problem of selecting the weights to learn the training set. The notion of closeness on the training set *T* is typically formalized through an error function of the form

$$\psi_T = \sum_{i=1}^{n} \|y_i - t_i\|^2 \tag{1}$$

where $y_i$ is the network output. Our target is to find a neural network $\eta$ such that the output $y_i = \eta(x_i, w)$ is close to the desired output $t_i$ for the input $x_i$ (w = strengths of synaptic connections). The error $\psi_T = \psi_T(w)$ is a function of $w$ because $y = \eta$ depends upon the parameters $w$ defining the selected network $\eta$. The objective function $\psi_T(w)$ for a neural network with many parameters defines a highly irregular surface with many local minima, large regions of little slope and symmetries. The common node functions (tanh, sigmoidal, logistic etc) are differentiable to arbitrary order through the chain rule of differentiation, which implies that the error is also differentiable to arbitrary order. Hence we are able to make a Taylor's series expansion in $w$ for $\psi_T$ [30]. We shall first discuss the algorithms for minimizing $\psi_T$ by assuming that we can truncate a Taylor's series expansion about a point $w^o$ that is possibly a local minimum. The gradient (first partial derivative) vector is represented by

$$g(w) = \nabla \psi_T \big|_w = \left[\frac{\partial \psi_T}{\partial w_i}\right]_w \tag{2}$$

The gradient vector points in the direction of steepest increase of $\psi_T$ and its negative points in the direction of steepest decrease. The second partial derivative also known as Hessian matrix is represented by *H*

$$H(w) = H_{ij}(w) = \nabla^2 \psi_T(w) = \frac{\partial^2 \psi_T(w)}{\partial w_i \partial w_j} \tag{3}$$

The Taylor's series for $\psi_T$, assumed twice continuously differentiable about $w^0$, can now be given as

$$\psi_T(w) = \psi_T(w^0) + g(w^0)^T(w - w^0)^T + \frac{1}{2}(w - w^0)^T H(w^0)(w - w^0) + O(\|w - w^0\|^2) \tag{4}$$

where $O(\delta)$ denotes a term that is of zero-order in small $\delta$ such that $\lim_{\delta \to 0} \frac{O(\delta)}{\delta} = 0$.

If for example there is continuous derivative at $w^0$, then the remainder term is of order $\|w - w^0\|^3$ and we can reduce (4) to the following quadratic model

$$m(w) = \psi_T(w^0) + g(w^0)^T(w - w^0) + \frac{1}{2}(w - w^0)^T H(w^0)(w - w^0) \tag{5}$$

Taking the gradient in the quadratic model of (5) yields

$$\nabla m = g(w^0) + H(w - w^0) \tag{6}$$

If we set the gradient $g=0$ and solving for the minimizing $w^*$ yields

$$w^* = w^0 - H^{-1}g \tag{7}$$

The model *m* can now be expressed in terms of minimum value of $w^*$ as

$$m(w^*) = m(w^0) + \frac{1}{2} g(w^0)^T H^{-1} g(w^0)$$

$$m(w) = m(w^*) + \frac{1}{2}(w - w^*)^T H(w^*)(w - w^*) \tag{8}$$

a result that follows from (5) by completing the square or recognizing that $g(w^*)=0$. Hence starting from any initial value of the weight vector, we can in the quadratic case move one step to the minimizing value when it exists. This is known as Newton's approach and can be used in the non-quadratic case where H is the Hessian and is positive definite.

## 2.1 Multiple Minima Problem in Neural Networks

A long recognized bane of analysis of the error surface and the performance of training algorithms is the presence of multiple stationary points, including multiple minima. Analysis of the behavior of training algorithms generally use the Taylor's series expansions discussed earlier, typically with the expansion about a local minimum $w^0$ However, the multiplicity of minima confuse the analysis because we need to be assured that we are converging to the same local minimum as used in the expansion. How likely are we to encounter a sizable number of local minima? Empirical experience with training algorithm shows that different initialization yield different resulting networks. Hence the issue of many minima is a real one. According to Auer et al [8], a single node network with $n$ training pairs and $R^d$ inputs could end up having $(\frac{n}{d})^d$ local minima. Hence not only multiple minima exist, but there may be huge numbers of them.

Different learning algorithms have their staunch proponents, who can always construct instances in which their algorithm perform better than most others. In practice, there are four types of optimization algorithms that are used to minimize $\Psi_T(w)$. The first three methods gradient descent, conjugate gradients and quasi-Newton are general optimization methods whose operation can be understood in the context of minimization of a quadratic error function. Although the error surface is surely not quadratic, for differentiable node functions it will be so in a sufficiently small neighborhood of a local minimum, and such an analysis provides information about the behavior of the training algorithm over the span of a few iterations and also as it approaches its goal. The fourth method of Levenberg and Marquardt is specifically adapted to minimization of an error function that arises from a squared error criterion of the form we are assuming. Backpropagation calculation of gradient can be adapted easily to provide the information about the Jacobian matrix $J$ needed for this method. A common feature of these training algorithms is the requirement of repeated efficient calculation of gradients.

### 2.1.1 Backpropagation Algorithm

Backpropagation provides an effective method for evaluating the gradient vector needed to implement the steepest descent, conjugate gradient, and quasi-Newton algorithms. BP differs from straightforward gradient calculations using the chain rule for differentiation in the way it organizes efficiently the gradient calculation for networks having more than one hidden layer [68]. BP iteratively selects a sequence of parameter vectors $\{w_k, k=1:T\}$ for a moderate value of running time T, with the goal of having $\{\Psi_T(w_k) = \Psi(k)\}$ converge to a small neighborhood of a good local minimum rather than the usually inaccessible global minimum [30].

$$\psi_T^* = min_{w \in W} \ \psi_T(w) \tag{9}$$

The simplest steepest descent algorithm uses the following weight update in the direction of $d_k=-g_k$ with a learning rate or step size $\alpha_k$.

$$w_{k+1} = w_k - \alpha_k g_k \tag{10}$$

A good choice $\alpha_k^*$ for the learning rate $\alpha_k$ for a given choice of descent direction $d_k$ is the one that minimizes $\psi_{(k+1)}$.

$$\alpha_k^* = arg \ min_\alpha \ \psi(w_k + \alpha d_k) \tag{11}$$

To carry out the minimization we use

$$\left. \frac{\partial \psi(w_{k+1})}{\partial \alpha} \right|_{\alpha=\alpha_k^*} = \left. \frac{\partial \psi(w_k + \alpha d_k)}{\partial \alpha} \right|_{\alpha=\alpha_k^*} = 0 \tag{12}$$

To evaluate this equation, note that

$$\frac{\partial \psi(w_k + \alpha d_k)}{\partial \alpha} = g_{k+1}^T d_k \tag{13}$$

and conclude that for optimal learning rate we must satisfy the orthogonality condition

$$g_{k+1}^T d_k = 0 \tag{14}$$

When the error function is not specified analytically, then its minimization along $d_k$ can be accomplished through a numerical line search for $\alpha_k$ or through numerical differentiation as noted herein. The line search avoids the problem of setting a fixed step size. Analysis of such algorithms often examine their behavior when the error function is truly a quadratic as given in (5) and (6). In the current notation,

$$g_{k+1} - g_k = \alpha_k H d_k \qquad (15)$$

Hence the optimality condition for the learning rate $\alpha_k$ derived from the orthogonality condition (14) becomes

$$\alpha_k^* = \frac{-d_k^T g_k}{d_k^T H d_k} \qquad (16)$$

When search directions are chosen via $d_k = -M_k g_k$, with $M_k$ symmetric, then the optimal learning rate is

$$\alpha_k^* = \frac{-g_k^T M_k g_k}{g_k^T M_k H M_k g_k} \qquad (17)$$

In the case of steepest descent for a quadratic error function, $M_k$ is the identity and

$$\alpha_k^* = \frac{-g_k^T g_k}{g_k^T H g_k} \qquad (18)$$

One can think of $\alpha_k^*$ as the reciprocal of an expected value of the eigen values $\{\lambda_i\}$ of the Hessian with probabilities determined by the squares of the coefficients of the gradient vector $g_k$ expanded in terms of the eigen vectors $\{e_i\}$ of the Hessian.

$$\frac{1}{\alpha_k^*} = \sum_{i=1}^{p} q_i \lambda_i, \; q_i = \frac{(g_k^T e_i)^2}{g_k^T g_k} \qquad (19)$$

The algorithm, even in the context of a truly quadratic error surface and with line search, suffers from greed. The successive directions do not generally support each other in that after two steps; say, the gradient is usually no longer orthogonal to the direction taken in the first step. In the quadratic case there exists a choice of learning rates that will drive the error to its absolute minimum in no more than $p+1$ steps where $p$ is the number of parameters [30]. To see this, note that

$$\psi(w) = \psi(w^*) + \frac{1}{2}(w - w^*)^T H(w - w^*) = \psi(w^*) + \tfrac{1}{2} g^T H^{-1} g \qquad (20)$$

It is easily verified that if $g_k = g(w_k)$ then

$$g_k = \left[ \prod_{1}^{k} (I - \alpha_j H) \right] g_0 \qquad (21)$$

Hence for $k \geq p$, we can achieve $g_k = 0$ simply by choosing $\alpha_1, \ldots \alpha_p$ any permutation of $1/\lambda_1 \ldots 1/\lambda_p$, the reciprocals of the eigen values of the Hessian $H$; the resulting product of matrices is a matrix that annihilates each of the $p$ eigen vectors and therefore any other vector that can be represented as their weighted sum. Of course, in practice, we do not know the eigen values and cannot implement this algorithm. However, this observation points out the distinction between optimality when one looks ahead only one step and optimality when one adopts a more distant horizon. Traditionally the step size is held at a constant value $\alpha_k = \alpha$. The simplicity of this approach is belied by the need to carefully select the learning rate. If the fixed step size is too large, then we leave ourselves open to overshooting the line search minimum, we may engage in oscillatory or divergent behavior, and we loose guarantees of monotone reduction of the error function $\psi_T$. If the step size is too small, then we may need a very large number of iterations $T$ before we achieve a sufficiently small value of the error function. A variation on the constant learning rate is to adopt a deterministic learning rate schedule that varies the learning rate dependant on the iteration number.

An ad hoc departure from steepest descent is to add memory to the recursion through momentum term. Now the change in parameter vector $w$ depends not only on the current gradient but also on the most recent change in parameter vector,

$$\Delta_{k+1} = w_{k+1} - w_k = \beta \Delta_k - \alpha_k g_k \; \text{ for } k \geq 0 \qquad (22)$$

what we gain is a high frequency smoothing effect through the momentum term. The change in parameter vector depends not only on the current gradient $g_{k-1}$ but also, in an exponentially decaying fashion (provided that $0 \leq \beta < 1$), on all previous gradients. If the succession of recent gradients has tended to alternate directions, then the sum will be relatively small and we will make only small changes in the parameter vector. This could occur if we are in the vicinity of a local minimum, successive changes would just serve to bounce us back and forth past the minimum. If, however, recent gradients tends to align, then we will make an even larger change in the parameter vector and thereby move more rapidly across a large region of descent and possibly across over a small region of ascent that screened off a deeper local minimum. Of course, if the learning rate $\alpha$ is well chosen, then successive gradients will tend to be orthogonal and a weighted sum will not cancel itself out.

### 2.1.2 Conjugate Gradient Algorithm

The motivation behind the conjugate gradient algorithm is that we wish to iteratively select search directions ($d_k$) that are non-interfering in the sense that successive minimizations along these directions do not undo the progress made by previous minimizations. The search direction is selected in such a way that at each iteratively selected parameter value $w_k$, the current gradient $g_k$ is orthogonal to all previous search directions $d_1,....d_{k-1}$. Hence, at any given step in the iteration, the error surface has a direction of steepest descent that is orthogonal to the linear subspace of parameters spanned by the prior search directions. Steepest descent merely assured us that the current gradient is orthogonal to the last search direction. If the error function $\{\Psi_T(w_k)\}$ is quadratic with positive definite Hessian $H$, choosing the search directions ($d_i$) to be $H$-conjugate and the $\alpha_i$ to satisfy (16) is equivalent to the orthogonality between the current gradient and the past search directions given by

$$(\forall i < k < p) d_i^T g_k = 0 \qquad (23)$$

it is easily verified that conjugate directions ($d_i$) also form a linearly independent set of directions in weight space [30]. If weight space has dimension p then of course there can be only $p$ linearly independent directions of vectors. Hence, it is possible to represent any point as a linear combination of no more than $p$ of the conjugate directions, and in particular if $w^*$ is the sought location of the minimum of the error function, then there exist coefficients such that

$$w^* - w_0 = \sum_{i=0}^{p-1} \alpha_i^* d_i \qquad (24)$$

Thus if the error surface is quadratic with a positive definite Hessian, then selecting $H$-conjugate search directions and learning rates according to (16) guarantees a minimum in no more than $p$ iterations. To be able to apply the method of conjugate gradients we must be able to determine such a set of directions and then solve for the correct coefficients. Conventional conjugate gradient algorithms use a line search to find the minimizing step and are initialized as follows

$$d_0 = g_0 \qquad (25)$$

introducing a scaling $\beta_k$ to be determined, and then iterate with the simple recursion

$$d_{k+1} = -g_{k+1} + \beta_k d_k \qquad (26)$$

According to the conjugacy condition in (23) and the recursion of (26) yield

$$d_k^T H d_{k+1} = 0 = d_k^T H(-g_{k+1} + \beta_k d_k) \qquad (27)$$

Solving yields the necessary condition that

$$\beta_k = \frac{d_k^T H g_{k+1}}{d_k^T H d_k} \qquad (28)$$

Induction can be established that this recursive definition of conjugate gradient search directions does indeed yield a fully conjugate set when the error function is quadratic, although the derivation of (28) only established that $d_k$ and $d_{k+1}$ are conjugate. A version of the conjugate gradient algorithm that does not require line searches was developed by Moller [61] and uses the finite difference method for estimating $Hd_k$. To monitor the sign of the product $d_k^T H d_k$, define $\delta$ by

$$\delta = d_k^T H d_k \qquad (29)$$

Moller introduces two new variables, $\lambda$ and $\bar{\lambda}$, to define an altered value of $\delta$, $\bar{\delta}$. These variables are charged with ensuring that $\bar{\delta} > 0$. Although this does not affect the error surface, and the Hessian with the quadratic approximation will still suggest there is a maximum along the search direction, the method produces a step size that shows good results in practice. $\bar{\delta}$ is defined as follows

$$\bar{\delta} = \delta + (\bar{\lambda} - \lambda) d_k^T d_k \qquad (30)$$

The requirement for $\bar{\delta} > 0$ gives a condition for $\bar{\lambda}$

$$\bar{\lambda} \succ \lambda - \frac{\delta}{d_k^T d_k} \qquad (31)$$

Moller then sets $\bar{\lambda} = 2(\lambda - \frac{\delta}{d_k^T d_k})$ to satisfy (2.31) and so ensures $\bar{\delta} > 0$. Substituting this in (2.30)

$$\bar{\delta} = -\delta + \lambda d_k^T d_k \qquad (32)$$

In order to get a good quadratic approximation of the error surface, a mechanism to raise or lower $\lambda$ is needed when the Hessian is positive definite. Detailed step-by-step description can be found in [61].

### 2.1.3 Quasi - Newton Algorithm

If the error surface is purely quadratic, as per (7) we can solve the minimizing weight vector in a single step through Newton's method. This solution requires knowledge of the Hessian and assumes it to be constant and positive definite. We need a solution method that can take into account the variation of $H(w)$ with $w$, knowing the fact that the error function is at best only approximately quadratic and removal from a local minimum the approximating quadratic surface is likely to have a Hessian that is not positive definite and the evaluation of true Hessian is computationally too expensive.

The quasi- Newton method addresses themselves to these tasks by first generalizing the iterative algorithm to the form

$$w_{k+1} = w_k - \alpha_k M_k g_k \qquad (33)$$

The choice of step size $\alpha_k$ to use with a search direction $d_k = M_k g_k$ is determined by an approximate line search, and use of line search is essential to the success of this method. The quasi-Newton method iteratively tracks the inverse of the Hessian without ever computing it directly. Let $q_k = g_{k+1} - g_k$, and consider the expansion for the gradient (quadratic case)

$$q_k = H_k(w_{k+1} - w_k) = H_k p_k \qquad (34)$$

If we can evaluate the difference of gradients for p linearly independent increments $p_0, \ldots p_{p-1}$ in the weight vectors, then we can solve for the Hessian (assumed constant). To do so, form the matrices $P$ with $i$th column the vector $p_{i-1}$ and $Q$ with $i$th column the vector $q_{i-1}$. Then we have the matrix equation

$$Q = H P \qquad (35)$$

which can be solved for the Hessian, when the columns of P are linearly independent, through

$$H = Q P^{-1} \qquad (36)$$

Thus from the increments in the gradient induced by the increments in the weight vectors as training proceeds, we have some hope of being able to track the Hessian. An approximation to the inverse $M$ of the Hessian is achieved by interchanging $q_k$ and $p_k$ in an approximation to the Hessian itself

$$M = P Q^{-1} \qquad (37)$$

Hence the information is available in the sequence of gradients that determine the $q_k$, and the sequence of search directions and learning rates that determine the $p_k$, to infer to the inverse of the Hessian, particularly if it is only slowly varying.

The Broyden-Fletcher-Goldfarb-Shanno (BFGS) quasi – Newton algorithm [25] implements the update for the approximate inverse M of the Hessian by

$$M_{k+1} = M_k + \left(1 + \frac{q_k^T M_k q_k}{q_k^T p_k}\right) \frac{p_k p_k^T}{p_k p_k^T} - \frac{p_k q_k^T M_k + M_k q_k p_k^T}{q_k^T p_k} \qquad (38)$$

This recursion is initialized by starting with a positive definite matrix such as the identity, $M_0 = I$. The Determination of the learning rates is critical, as was the case for the method of conjugate directions. Quasi-Newton methods enjoy asymptotically more rapid convergence than that of steepest descent or conjugate gradient methods.

### 2.1.4 Levenberg-Marquardt algorithm

The Levenberg-Marquardt (LM) algorithm [25] exploits the fact that the error function is a sum of squares as given in (1). Introduce the following notation for the error vector and its Jacobian with respect to the network parameters w

$$J = J_{ij} = \frac{\partial e_j}{\partial w_i}, i = 1 : p, j = 1 : n \qquad (39)$$

The Jacobian matrix is a large p × n matrix, all of whose elements are calculated directly by backpropagation technique as presented in Section 2.1.1. The $p$ dimensional gradient $g$ for the quadratic error function can be expressed as

$$g(w) = \sum_{i=1}^{n} e_i \nabla e_i(w) = Je$$

and the Hessian matrix by

$$H = H_{ij} = \frac{\partial^2 \psi_T}{\partial w_i \partial w_j} = \frac{1}{2} \sum_{k=1}^{n} \frac{\partial^2 e_k^2}{\partial w_i \partial w_j} = \sum_{k=1}^{n} \left( e_k \frac{\partial^2 e_k}{\partial w_i \partial w_j} + \frac{\partial e_k \partial e_k}{\partial w_i \partial w_j} \right)$$

$$= \sum_{k=1}^{n} \left( e_k \frac{\partial^2 e_k}{\partial w_i \partial w_j} + J_{ik} J_{jk} \right) \qquad (40)$$

Hence defining $D = \sum_{i=1}^{n} e_i \nabla^2 e_i$ yields the expression

H (w) = JJ$^T$ + D  (41)

The key to the LM algorithm is to approximate this expression for the Hessian by replacing the matrix $D$ involving second derivatives by the much simpler positively scaled unit matrix $\in I$. The LM is a descent algorithm using this approximation in the form

$$M_k = \left[JJ^T + \in I\right]^{-1}, w_{k+1} = w_k - \alpha_k M_k g(w_k) \qquad (42)$$

Successful use of LM requires approximate line search to determine the rate $\alpha_k$. The matrix JJ$^T$ is automatically symmetric and non-negative definite. The typically large size of J may necessitate careful memory management in evaluating the product JJ$^T$. Hence any positive $\in$ will ensure that $M_k$ is positive definite, as required by the descent condition. The performance of the algorithm thus depends on the choice of $\in$.

When the scalar $\in$ is zero, this is just Newton's method, using the approximate Hessian matrix. When $\in$ is large, this becomes gradient descent with a small step size. As Newton's method is more accurate, $\in$ is decreased after each successful step (reduction in performance function) and is increased only when a tentative step would increase the performance function. By doing this, the performance function will always be reduced at each iteration of the algorithm [12].

## 2.2. Designing Artificial Neural Networks

The error surface of very small networks has been characterized previously. However, practical networks often contain hundreds of weights and in general, theoretical and empirical results on small networks do not scale up to large networks. To investigate the empirical performance with the different learning algorithms on different architectures and node transfer functions, we have choosen 3 famous chaotic time series benchmarks so that **a)** we know the best solution, **b)** can carefully control various parameters and **c)** know the effect of the different learning algorithms namely backpropagation (BP), scaled conjugate gradient (SCG), quasi-Newton algorithm (QNA) and Levenberg Marquardt algorithm (LM).

We also report some experimentation results related to convergence speed and generalization performance of the four different neural network-learning algorithms discussed in Section 2.1. Performances of the different learning algorithms were evaluated when the activation functions and architectures were changed.

We used a feedforward neural network with 1 hidden layer and the numbers of hidden neurons were varied (14,16,18,20,24) and the speed of convergence and generalization error for each of the four learning algorithms was observed. The effect of node activation functions, log-sigmoidal activation function (LSAF) and tanh-sigmoidal activation function (TSAF), keeping 24 hidden neurons for the four learning algorithms was also studied. Computational complexities of the different learning algorithms were also noted during each event. The experiments were replicated 3 times each with a different starting condition (random weights) and the worst errors were reported. No stopping criterion, and no method of controlling generalization is used other than the maximum number of updates (epochs). All networks were trained for an identical number of stochastic updates (2500 epochs). We used the following three chaotic time series:

### a) Waste Water Flow Prediction

The problem is to predict the wastewater flow into a sewage plant [46]. The water flow was measured every hour. It is important to be able to predict the volume of flow $f(t+1)$ as the collecting tank has a limited capacity and a sudden increase in flow will cause to overflow excess water. The water flow prediction is to assist an adaptive online controller. The data set is represented as $[f(t), f(t-1), a(t), b(t), f(t+1)]$ where $f(t), f(t-1)$ and $f(t+1)$ are the water flows at time $t, t-1,$ and $t+1$ (hours) respectively. $a(t)$ and $b(t)$ are the moving averages for 12 hours and 24 hours. The time series consists of 475 data points. The first 240 data sets were used for training and remaining data for testing.

### b) Mackey-Glass Chaotic Time Series

The Mackey-Glass differential equation [53] is a chaotic time series for some values of the parameters $x(0)$ and $\tau$.

$$\frac{dx(t)}{dt} = \frac{0.2x(t-\tau)}{1 + x^{10}(t-\tau)} - 0.1\, x(t). \qquad (43)$$

We used the value $x(t-18), x(t-12), x(t-6), x(t)$ to predict $x(t+6)$. Fourth order Runge-Kutta method was used to generate 1000 data series. The time step used in the method is 0.1 and initial condition were $x(0)=1.2$, $\tau=17$, $x(t)=0$ for $t<0$. First 500 data sets were used for training and remaining data for testing.

### c) Gas Furnace Time Series Data

This time series was used to predict the $CO_2$ (carbon dioxide) concentration $y(t+1)$ [17]. In a gas furnace system, air and methane are combined to form a mixture of gases containing $CO_2$. Air fed into the gas furnace is kept constant, while the methane feed rate $u(t)$ can be varied in any desired manner. After that, the resulting $CO_2$ concentration $y(t)$ is measured in the exhaust gases at the outlet of the furnace. Data is represented as $[u(t), y(t), y(t+1)]$. The time series consists of 292 pairs of observation and 50% of data was used for training and remaining for testing.

### 2.2.1 Simulation Results Using ANNs

Results for four different learning algorithms for different architectures, node transfer functions for the three different time series are presented in the following sections.

### 2.2.1.1 Network Architecture

This section investigates the training and generalization behavior of the networks when the architecture of the neural network was changed. The same architecture was used for the three different time series for the four learning algorithms using same node transfer function. Tables 1–3 summarize the empirical results of training and generalization.

Table 1. Training and test performance for Mackey Glass Series for different architectures

| Mackey Glass Time Series | | | |
| --- | --- | --- | --- |
| Learning algorithm | Hidden Neurons | Root Mean Squared Error | |
| | | Training data | Test data |
| BP | 14 | 0.0890 | 0.0880 |
| | 16 | 0.0824 | 0.0860 |
| | 18 | 0.0764 | 0.0750 |
| | 20 | 0.0452 | 0.0442 |
| | 24 | 0.0439 | 0.0437 |
| SCG | 14 | 0.0040 | 0.0051 |
| | 16 | 0.0053 | 0.0052 |
| | 18 | 0.0066 | 0.0067 |
| | 20 | 0.0058 | 0.0058 |
| | 24 | 0.0045 | 0.0045 |
| QNA | 14 | 0.0041 | 0.0040 |
| | 16 | 0.0031 | 0.0030 |
| | 18 | 0.0035 | 0.0036 |
| | 20 | 0.0038 | 0.0038 |
| | 24 | 0.0034 | 0.0036 |
| LM | 14 | 0.0016 | 0.0016 |
| | 16 | 0.0015 | 0.0015 |
| | 18 | 0.0015 | 0.0015 |
| | 20 | 0.0010 | 0.0011 |
| | 24 | 0.0009 | 0.0009 |

Table.2. Training and test performance for gas furnace time series for different architectures

| Gas Furnace Time Series | | | |
| --- | --- | --- | --- |
| Learning algorithm | Hidden Neurons | Root Mean Squared Error | |
| | | Training data | Test data |
| BP | 14 | 0.0670 | 0.1291 |
| | 16 | 0.0835 | 0.1056 |
| | 18 | 0.0716 | 0.0766 |
| | 20 | 0.0800 | 0.0950 |
| | 24 | 0.0663 | 0.0970 |
| SCG | 14 | 0.0160 | 0.0331 |
| | 16 | 0.0157 | 0.0330 |
| | 18 | 0.0165 | 0.0330 |
| | 20 | 0.0158 | 0.0361 |
| | 24 | 0.0153 | 0.0367 |
| QNA | 14 | 0.0137 | 0.0529 |
| | 16 | 0.0133 | 0.0465 |
| | 18 | 0.0133 | 0.0376 |
| | 20 | 0.0136 | 0.0410 |
| | 24 | 0.0128 | 0.0516 |
| LM | 14 | 0.0118 | 0.0450 |
| | 16 | 0.0140 | 0.0971 |
| | 18 | 0.0116 | 0.1080 |
| | 20 | 0.0100 | 0.1880 |
| | 24 | 0.0100 | 0.1856 |

## 2.2.1.2 Node transfer functions

This section investigates the effect of different node transfer functions on training and generalization performance for the four learning algorithms. To compare empirically we maintained the same architecture and only changing he node transfer functions and learning algorithms. All the networks were randomly initialized and trained for 2500 epochs. Tables 4 – 6 summarizes the empirical results of training and generalization for the two node transfer functions, tanh-sigmoidal activation function (TSAF) and log-sigmoidal activation function (LSAF), when the architecture was fixed with 24 hidden neurons.

**Table 3.** Training and test performance for wastewater flow series for different architectures

| Wastewater Time Series | | | |
|---|---|---|---|
| **Learning algorithm** | **Hidden Neurons** | **Root Mean Squared Error** | |
| | | **Training data** | **Test data** |
| BP | 14 | 0.1269 | 0.1340 |
| | 16 | 0.1184 | 0.1360 |
| | 18 | 0.1182 | 0.1350 |
| | 20 | 0.1221 | 0.1370 |
| | 24 | 0.1169 | 0.1412 |
| SCG | 14 | 0.0459 | 0.0900 |
| | 16 | 0.0428 | 0.1130 |
| | 18 | 0.0425 | 0.1130 |
| | 20 | 0.0423 | 0.1626 |
| | 24 | 0.0400 | 0.0920 |
| QNA | 14 | 0.0423 | 0.1271 |
| | 16 | 0.0367 | 0.1369 |
| | 18 | 0.0363 | 0.1360 |
| | 20 | 0.0339 | 0.1450 |
| | 24 | 0.0316 | 0.2620 |
| LM | 14 | 0.0364 | 0.0950 |
| | 16 | 0.0303 | 0.1631 |
| | 18 | 0.0314 | 0.1800 |
| | 20 | 0.0259 | 0.1314 |
| | 24 | 0.0244 | 0.1560 |

**Table.4.** Mackey Glass time series: Training and generalization performance for different activation functions

| Time series | Learning algorithm | Activation function | Root Mean Squared Error | |
|---|---|---|---|---|
| | | | **Training** | **Test** |
| Mackey Glass | BP | TSAF | 0.0439 | 0.0437 |
| | | LSAF | 0.0970 | 0.0950 |
| | SCG | TSAF | 0.0045 | 0.0045 |
| | | LSAF | 0.0076 | 0.0074 |
| | QNA | TSAF | 0.0033 | 0.0034 |
| | | LSAF | 0.0029 | 0.0029 |
| | LM | TSAF | 0.0009 | 0.0009 |
| | | LSAF | 0.0009 | 0.0010 |

### 2.2.1.3 Computational Complexity of Learning algorithms

This section investigates the computational complexity of the different learning algorithms when the architecture of the hidden layer is varied using tanh-sigmoidal activation function. The networks were randomly initialized and trained for 2500 epochs using the different learning algorithms. Table 7 summarizes the empirical values of the computational load for the different learning methods for the three different time series.

**Table 5.** Gas furnace series: Training and generalization performance for different activation functions

| Time series | Learning algorithm | Activation function | Root Mean Squared Error | |
|---|---|---|---|---|
| | | | Training | Test |
| Gas furnace | BP | TSAF | 0.0663 | 0.0970 |
| | | LSAF | 0.0940 | 0.1025 |
| | SCG | TSAF | 0.0153 | 0.0367 |
| | | LSAF | 0.0162 | 0.0367 |
| | QNA | TSAF | 0.0128 | 0.0516 |
| | | LSAF | 0.0137 | 0.0420 |
| | LM | TSAF | 0.0100 | 0.1856 |
| | | LSAF | 0.0089 | 0.1009 |

**Table 6.** Waste water time series: Training and generalization performance for different activation functions

| Time series | Learning algorithm | Activation function | Root Mean Squared Error | |
|---|---|---|---|---|
| | | | Training | Test |
| Wastewater | BP | TSAF | 0.1169 | 0.1412 |
| | | LSAF | 0.0156 | 0.1600 |
| | SCG | TSAF | 0.0400 | 0.0920 |
| | | LSAF | 0.0420 | 0.0820 |
| | QNA | TSAF | 0.0316 | 0.4600 |
| | | LSAF | 0.0256 | 0.2110 |
| | LM | TSAF | 0.0244 | 0.1560 |
| | | LSAF | 0.2160 | 0.1770 |

**Table 7.** Approximate computational load for the different time series using the different training algorithms

| Learning algorithm | Hidden Neurons | Computational Load (billion flops) | | |
|---|---|---|---|---|
| | | Mackey Glass | Gas Furnace | Waste water |
| BP | 14 | 0.625 | 0.142 | 0.301 |
| | 16 | 0.713 | 0.305 | 0.645 |
| | 18 | 0.800 | 0.488 | 0.880 |
| | 20 | 0.888 | 0.690 | 1.460 |
| | 24 | 1.064 | 0.932 | 1.970 |
| SCG | 14 | 1.256 | 0.286 | 0.604 |
| | 16 | 1.429 | 0.326 | 0.689 |
| | 18 | 1.605 | 0.366 | 0.774 |
| | 20 | 1.781 | 0.406 | 0.859 |
| | 24 | 2.133 | 0.486 | 1.029 |
| QNA | 14 | 2.570 | 0.679 | 1.910 |
| | 16 | 3.319 | 0.8899 | 2.582 |
| | 18 | 4.221 | 0.9000 | 3.388 |
| | 20 | 5.313 | 1.131 | 4.384 |
| | 24 | 7.989 | 2.193 | 6.925 |
| LM | 14 | 29.40 | 3.930 | 12.46 |
| | 16 | 57.51 | 8.355 | 27.72 |
| | 18 | 93.29 | 14.03 | 93.79 |
| | 20 | 137.83 | 21.10 | 118.53 |
| | 24 | 203.10 | 31.83 | 175.22 |

## 2.3. Discussion of Results Obtained

In this Section we would like to evaluate and summarize the results of the various experimentations mentioned in Section 2.2.1. For Mackey Glass series (Table 1) all the 4 learning algorithms tend to generalize well as the hidden neurons were increased. However the generalization was better when the hidden neurons were using TSAF. LM showed the fastest convergence regardless of architecture and node activation function. However, the computational complexity of LM algorithm is very amazing as depicted in Table 7. For Mackey glass series (with 14 hidden neurons), when BP was using 0.625 billion flops, LM technique required 29.4 billion flops. When the hidden neurons were increased to 24, BP used 1.064 billion flops and LM's share jumped to 203.10 billion flops. LM gave the lowest generalization RMSE of 0.0009 with 24 hidden neurons.

As shown in Table 2, for gas furnace series the generalization performance were entirely different for the different learning algorithms. BP gave the best generalization RMSE of 0.0766 with 18 hidden neurons. RMSE for SCG, QNA and LM were 0.0330 (16 neurons), 0.0376 (18 neurons) and 0.045 (14 neurons) respectively. As depicted in Figures 9 and 10 the node transfer function also has an effect on the training speed and generalization performance. LM algorithm converged much faster and gave a better generalization performance when the node transfer function was changed to LSAF (Refer to Figure 10(b)).

Waste water prediction series also showed a different generalization performance when the architecture was changed for the different learning algorithms (Refer to Table 3). BP's best generalization RMSE was 0.135 with 18 hidden neurons using TSAF and that of SCG, QNA and LM were 0.0900, 0.1271 and 0.095 with 14 neurons each respectively.

LM algorithm converged much faster and gave a better generalization performance when the node transfer function was changed to LSAF (Refer to Figure 12(b)).

In spite of computational complexity, LM performed well for Mackey Glass series. For gas furnace and waste water prediction SCG algorithm performed better. However the speed of convergence of LM in all the three cases is worth noting. This leads us to the following questions:

- What is the optimal architecture (number of neurons and hidden layers) for a given problem?
- What node transfer function(s) should one choose?
- What is the optimal learning algorithm and its parameters?

From the above discussion it is clear that the selection of the topology of a network and the best learning algorithm and its parameters is a tedious task for designing an optimal artificial neural network, which is smaller, faster and with a better generalization performance. Evolutionary algorithm is an adaptive search technique based on the principles and mechanisms of natural selection and survival of the fittest from natural evolution [31]. The interest in evolutionary search procedures for designing neural network topology has been growing in recent years as they can evolve towards the optimal architecture without outside interference, thus eliminating the tedious trial and error work of manually finding an optimal network.

## 3. Evolutionary Algorithms (EA)

EAs are population based adaptive methods, which may be used to solve optimization problems, based on the genetic processes of biological organisms [31] [32]. Over many generations, natural populations evolve according to the principles of natural selection and "Survival of the Fittest", first clearly stated by Charles Darwin in "On the Origin of Species". By mimicking this process, EAs are able to "evolve" solutions to real world problems, if they have been suitably encoded. The procedure may be written as the difference equation [31]:

$$x[t + 1] = s(v(x[t]))  \qquad (44)$$

where $x(t)$ is the population at time $t$, $v$ is a random operator, and $s$ is the selection operator.

## 3.2 Evolutionary Artificial Neural Networks (EANN)

Many of the conventional ANNs now being designed are statistically quite accurate but they still leave a bad taste with users who expect computers to solve their problems accurately. The important drawback is that the designer has to specify the number of neurons, their distribution over several layers and interconnection between them. Several methods have been proposed to automatically construct ANNs for reduction in network complexity that is to determine the appropriate number of hidden units, layers, etc. Topological optimization algorithms such as Extentron [9], Upstart [35], Pruning [63] [75] and Cascade Correlation [29] etc. got its own limitations.

The interest in evolutionary search procedures for designing ANN architecture has been growing in recent years as they can evolve towards the optimal architecture without outside interference, thus eliminating the tedious trial and error work of manually finding an optimal network [2] [6] [7] [14] [18] [19] [20] [21] [37] [50] [60] [69] [70] [77] [81] [83] [84] [85]. The advantage of the automatic design over the manual design becomes clearer as the complexity of ANN increases. EANNs provide a general framework for investigating various aspects of simulated evolution and learning [10] [14] [15] [50] [52].

### 3.2.1 General Framework for EANNs

In EANN's evolution can be introduced at various levels. At the lowest level, evolution can be introduced into weight training, where ANN weights are evolved. At the next higher level, evolution can be introduced into neural network architecture adaptation, where the architecture (number of hidden layers, no of hidden neurons and node transfer functions) is evolved. At the highest level, evolution can be introduced into the learning mechanism. A general framework of EANNs which includes the above three levels of evolution is given in Figure 1 [2] [6].

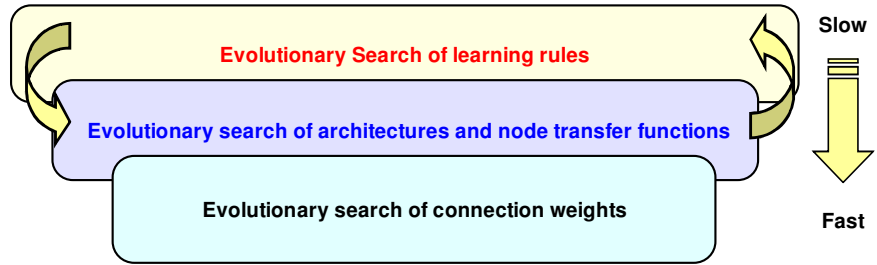

**Figure 1.** A General Framework for EANNs

From the point of view of engineering, the decision on the level of evolution depends on what kind of prior knowledge is available. If there is more prior knowledge about EANN's architectures than that about their learning rules or a particular class of architectures is pursued, it is better to implement the evolution of architectures at the highest level because such knowledge can be used to reduce the search space and the lower level evolution of learning rules can be more biased towards this kind of architectures. On the other hand, the evolution of learning rules should be at the highest level if there is more prior knowledge about them available or there is a special interest in certain type of learning rules.

### 3.2.1.1 Evolutionary Search of Connection weights

The shortcomings of the BP algorithm mentioned in Section 2.1 could be overcome if the training process is formulated as a global search of connection weights towards an optimal set defined by the evolutionary algorithm.. Optimal connection weights can be formulated as a global search problem wherein the architecture of the neural network is pre-defined and fixed during the evolution. Connection weights may be represented as binary strings represented by a certain length. The whole network is encoded by concatenation of all the connection weights of the network in the chromosome. A heuristic concerning the order of the concatenation is to put connection weights to the same node together. Figure 2 illustrates the binary representation of connection weights wherein each weight is represented by 4 bits.

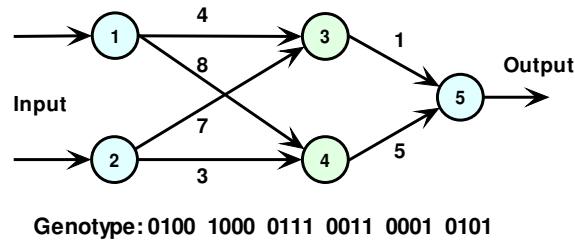

**Figure 2.** Connection weight chromosome encoding using binary representation

Real numbers have been proposed to represent connection weights directly [66]. A representation of the ANN could be (2.0, 6.0, 5.0, 1.0, 4.0, 10.0). However proper genetic operators are to be chosen depending upon the representation used.

Evolutionary Search of connection weights can be formulated as follows:

1) *Generate an initial population of N weight chromosomes. Evaluate the fitness of each EANN depending on the problem.*

2) *Depending on the fitness and using suitable selection methods reproduce a number of children for each individual in the current generation.*

3) *Apply genetic operators to each child individual generated above and obtain the next generation.*

4) *Check whether the network has achieved the required error rate or the specified number of generations has been reached. Go to Step 2.*

5) *End*

While gradient based techniques are very much dependant on the initial setting of weights, the proposed algorithm can be considered generally much less sensitive to initial conditions. When compared to any gradient descent or second order optimization technique that can only find local optimum in a neighborhood of the initial solution, evolutionary algorithms always try to search for a global optimal solution. Performance by using the above approach will directly depend on the problem.

### 3.2.1.2 Evolutionary Search of Architectures

Evolutionary architecture adaptation can be achieved by constructive and destructive algorithms. Constructive algorithms, which add complexity to the network starting from a very simple architecture until the entire network is able to learn the task [35] [56] [59]. Destructive algorithms start with large architectures and remove nodes and interconnections until the ANN is no longer able to perform its task [63] [75]. Then the last removal is undone. Figure 3 demonstrates how typical neural network architecture could be directly encoded and how the genotype is represented. For an optimal network, the required node transfer function (Gaussian, sigmoidal, etc.) can be formulated as a global search problem, which is evolved simultaneously with the search for architectures [51].

To minimize the size of the genotype string and improve scalability, when priori knowledge of the architecture is known it will be efficient to use some indirect coding (high level) schemes. For example, if two neighboring layers are fully connected then the architecture can be coded by simply using the number of layers and nodes. The blueprint representation is a popular indirect coding scheme where it assumes architecture consists of various segments or areas. Each segment or area will define a set of neurons, their spatial arrangement and their efferent connectivity. Several high level coding schemes like graph generation system [49], Symbiotic Adaptive Neuro-Evolution (SANE) [62] [65], marker based genetic coding [36], L-systems [13], cellular encoding [38], fractal representation [58] etc are some of the rugged techniques.

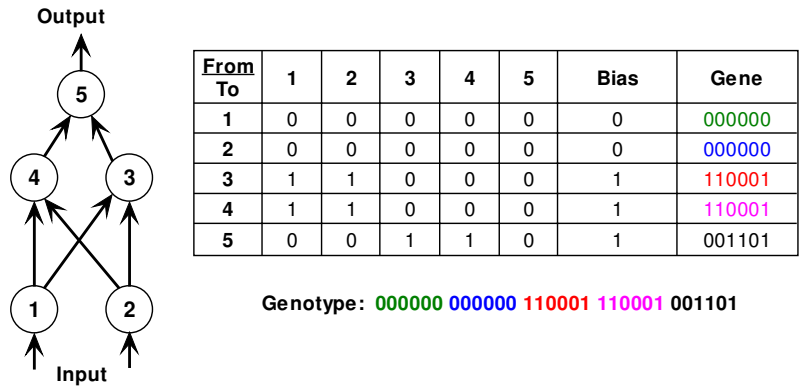

**Figure 3.** Architecture chromosome using binary coding

Global search of transfer function and the connectivity of the ANN using evolutionary algorithms can be formulated as follows

1) *The evolution of architectures has to be implemented such that the evolution of weight chromosomes are evolved at a faster rate i.e. for every architecture chromosome, there will be several weight chromosomes evolving at a faster time scale*

2) *Generate an initial population of N architecture chromosomes. Evaluate the fitness of each EANN depending on the problem.*

3) *Depending on the fitness and using suitable selection methods reproduce a number of children for each individual in the current generation.*

4) *Apply genetic operators to each child individual generated above and obtain the next generation.*

5) *Check whether the network has achieved the required error rate or the specified number of generations has been reached. Go to Step 3.*

6) *End*

### 3.2.1.3 Evolutionary Search of Learning Rules

For the neural network to be fully optimal the learning rules are to be adapted dynamically according to its architecture and the given problem. Deciding the learning rate and momentum can be considered as the first attempt of learning rules [48]. The basic learning rule can be generalized by the function

$$\Delta w(t) = \sum_{k=1}^{n} \sum_{i_1,i_2,\ldots,i_k=1}^{n} \left( \theta_{i_1,i_2,\ldots,i_k} \prod_{j=1}^{k} x_{i_j}(t-1) \right) \qquad (45)$$

Where $t$ is the time, $\Delta w$ is the weight change, $x_1, x_2, \ldots x_n$ are local variables and the $\theta$'s are the real values coefficients which will be determined by the global search algorithm. In the above equation different values of $\theta$'s determine different learning rules. The above equation is arrived based on the assumption that the same rule is applicable at every node of the network and the weight updating is only dependent on the input/output activations and the connection weights on a particular node. Genotypes ($\theta$'s) can be encoded as real-valued coefficients and the global search for learning rules using the hybrid algorithm can be formulated as follows:

1.  *The evolution of learning rules has to be implemented such that the evolution of architecture chromosomes are evolved at a faster rate i.e. for every learning rule chromosome, there will be several architecture chromosomes evolving at a faster time scale*

2.  *Generate an initial population of N learning rules. Evaluate the fitness of each EANN depending on the problem.*

3.  *Depending on the fitness and using suitable selection methods reproduce a number of children for each individual in the current generation.*

4.  *Apply genetic operators to each child individual generated above and obtain the next generation.*

5.  *Check whether the network has achieved the required error rate or the specified number of generations has been reached. Go to Step 3.*

6.  *End*

Several researches have been going on about how to formulate different optimal learning rules [4] [6] [11] [33] [82]. The adaptive adjustment of BP algorithm's parameters, such as the learning rate and momentum, through evolution could be considered as the first attempt of the evolution of learning rules [40]. Chalmers [23] defined the form of learning rules as a linear function of four local variables and their six pair wise products [11] [33]. Global optimization of neural network has been widely addressed using several other techniques [22] [28] [34] [64] [71] [72] [73] [74] [86]. Sexton et al [72] used simulated annealing algorithm for optimization of learning. For optimization of the neural network learning, in many cases a pre-defined architecture was used and in a few cases architectures were evolved together. No work has been reported to the best of our knowledge, where the network is fully automated (interaction of the different evolutionary search mechanisms) using the generic framework mentioned in Section 3.2. Many a times, the search space is narrowed down by pre-defined architecture, node transfer functions and learning rules.

## 3.3 Meta Learning Evolutionary Artificial Neural Networks (MLEANN)

Experimental evidence had indicated cases where evolutionary algorithms are inefficient at fine tuning solutions, but better at finding global basins of attraction [2] [82] [43] [80]. The efficiency of evolutionary training can be improved significantly by incorporating a local search procedure into the evolution. Evolutionary algorithms are used to first locate a good region in the space and then a local search procedure is used to find a near optimal solution in this region. It is interesting to consider finding good initial weights as locating a good region in the space. Defining that the basin of attraction of a local minimum is composed of all the points, sets of weights in this case, which can converge to the local minimum through a local search algorithm, then a global minimum can easily be found by the local search algorithm if the evolutionary algorithm can locate any point, i.e, a set of initial weights, in the basin of attraction of the global minimum. Referring to Figure 4, $G_1$ and $G_2$ could be considered as the initial weights as located by the evolutionary search and $W_A$ and $W_B$ the corresponding final weights fine-tuned by the meta-learning technique.

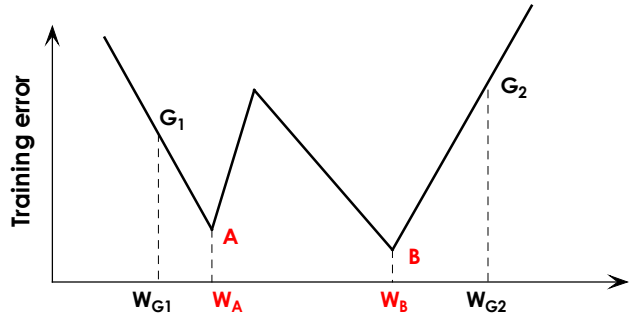

**Figure 4.** Fine tuning of weights using meta-learning

Figure 5 illustrates the general interaction mechanism with the learning mechanism of the EANN evolving at the highest level on the slowest time scale. All the randomly generated architecture of the initial population are trained by four different learning algorithms (backpropagation-BP, scaled conjugate gradient-SCG, quasi-Newton algorithm-QNA and Levenberg-Marquardt-LM) and evolved in a parallel environment. Parameters controlling the performance of the learning algorithm will be adapted (example, learning rate and momentum for BP) according to the problem [2] [6]. Figure 6 depicts the basic algorithm of proposed meta-learning EANN. Architecture of the chromosome is depicted in Figure 7.

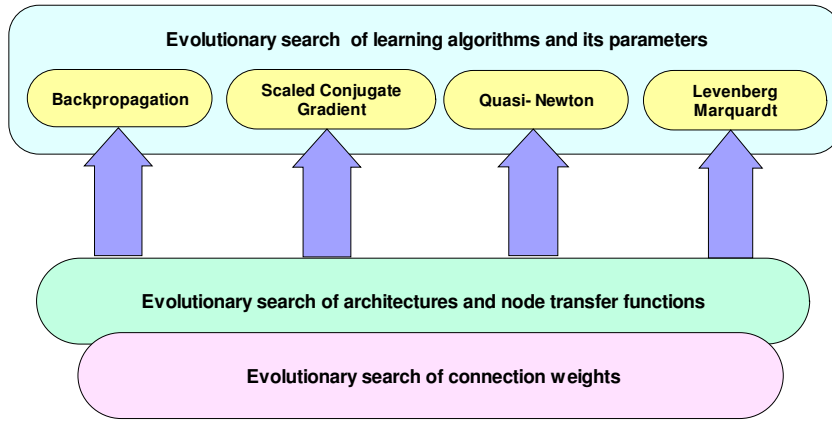

**Figure 5.** Interaction of various evolutionary search mechanisms

1. *Set t=0 and randomly generate an initial population of neural networks with architectures, node transfer functions and connection weights assigned at random.*
2. *In a parallel mode, evaluate fitness of each ANN using BP/SCG/QNA and LM*
3. *Based on fitness value, select parents for reproduction*
4. *Apply mutation to the parents and produce offspring (s) for next generation. Refill the population back to the defined size.*
5. *Repeat step 2*
6. *STOP when the required solution is found or number of iterations has*

**Figure 6**. Meta-learning algorithm for EANNs

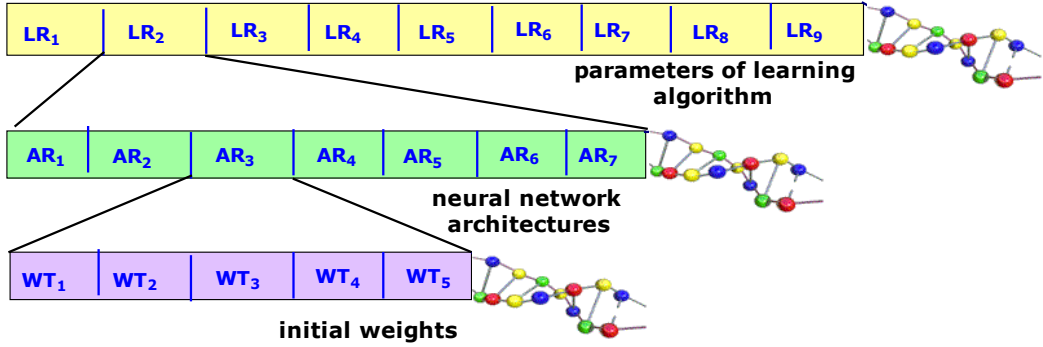

**Figure 7.** Chromosome representation of the proposed MLEANN framework

### 3.3.1 MLEANN: Experiment Setup

We have applied the proposed meta learning framework to the three-time series prediction problems discussed in Section 2.2. For performance comparison, we used the same set of training and test data that were used for experimentations with conventional design of neural networks. For performance evaluation, the parameters used in our experiments were set to be the same for all the 3 problems. Fitness value is calculated based on the RMSE achieved on the test set. In this experiment, we have considered the best-evolved neural network as the best individual of the last generation. As the learning process is evolved separately, user has the option to pick the best neural network (e.g. less RMSE or less computational expensive etc.) among the four learning algorithms. All the genotypes were represented using binary coding and the initial populations were randomly generated based on the following parameters shown in Table 8. The parameter settings, which were evolved for the different learning algorithms, are illustrated in Table 9. The parameter settings mentioned in Table 8 and 9 were finalized after a few trail and error approaches. We also investigated the performance of the proposed method with a restriction of architecture (no of hidden neurons). We set a maximum number of 4 hidden neurons and evaluated the learning performance. The experiments were repeated three times and the worst RMSE values are reported.

### 3.3.2 MLEANN: Experimentation Results

Table 10 displays empirical values of RMSE on test data for the three time series problems without architecture restriction. For comparison purposes, test set RMSE values using conventional design techniques are also presented in Table 10 (adapted from Table 1,2 and 3). Table 11 illustrates the RMSE values on training/test set data using the meta-learning technique when the architecture restriction was imposed. Run times for the two different experimentations are also presented.

**Table 8.** Parameters used for evolutionary design of artificial neural networks

| Population size | 40 |
|---|---|
| Maximum no of generations | 40 |
| Number of hidden nodes | • Experiment 1: 5-16 hidden nodes <br> • Experiment 2: maximum 4 neurons |
| Activation functions | tanh *(T)*, logistic *(L)*, sigmoidal *(S)*, tanh-sigmoidal *(T\*)*, log-sigmoidal *(L\*)* |
| Output neuron | linear |
| Training epochs | 500 |
| Initialization of weights | +/- 0.3 |
| Ranked based selection | 0.50 |
| Elitism | 5 % |
| Mutation rate | 0.40 |

**Table 9.** Parameters settings of the learning algorithms

| Learning algorithm | Parameter | Setting |
|---|---|---|
| Backpropagation | Learning rate | 0.25-0.05 |
| | Momentum | 0.25-0.05 |
| Scaled conjugate gradient algorithm | Change in weight for second derivative approximation | 0 - 0.0001 |
| | Regulating the indefiniteness of the Hessian | 0 – 1.0 E-06 |
| Quasi-Newton algorithm | Step lengths | 1.0E-06 – 100 |
| | Limits on step sizes | 0.1 – 0.6 |
| | Scale factor to determine performance | 0.001 – 0.003 |
| | Scale factor to determine step size. | 0.1 - 0.4 |
| Levenberg Marquardt | Learning rate | 0.001 – 0.02 |

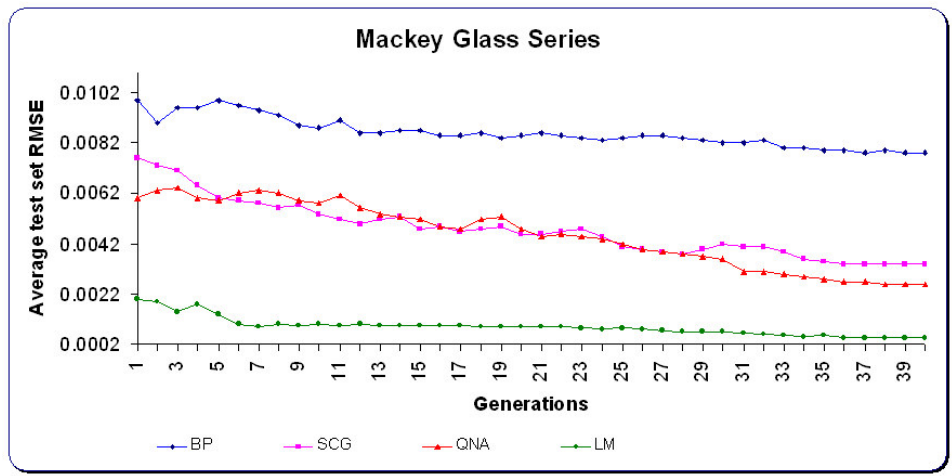

**Figure 8.** Mackey Glass time series: Average test set RSME values during the 40 generations and meta-learning

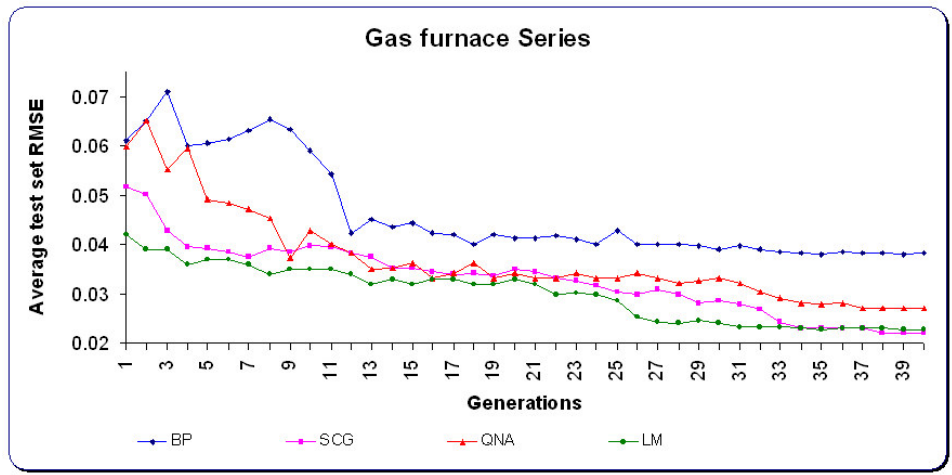

**Figure 9.** Gas furnace time series: Average test set RSME values during the 40 generations and meta-learning

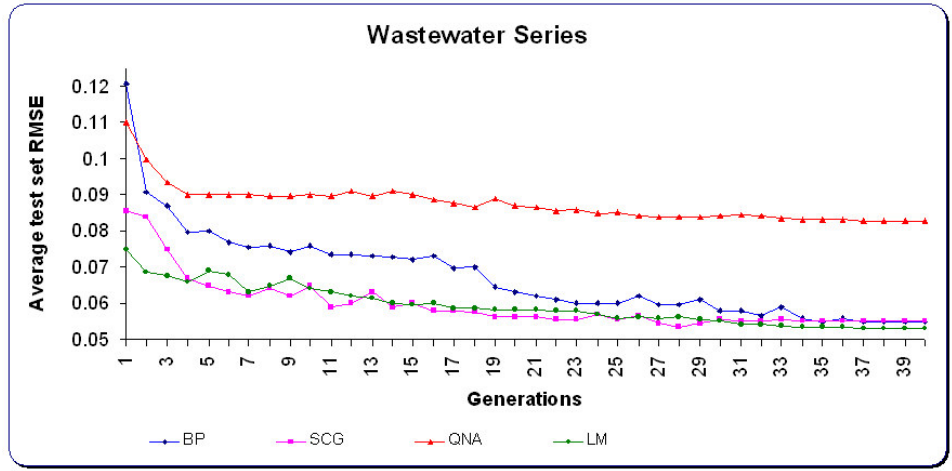

**Figure 10.** Wastewater time series: Average test set RSME values during the 40 generations and meta-learning

**Table 10.** Performance comparison between MLEANN (without architecture restriction) and ANN

| Time series | Learn Algo. | EANN RMSE Training | EANN RMSE Test | Architecture | ANN RMSE | ANN Architecture |
|---|---|---|---|---|---|---|
| Mackey Glass | BP | 0.0072 | 0.0077 | 7 T, 3 L | 0.0437 | 24 T* |
| | SCG | 0.0030 | 0.0031 | 11 T | 0.0045 | 24 T* |
| | QNA | 0.0024 | 0.0027 | 6 T, 4 T* | 0.0034 | 24 T* |
| | LM | 0.0004 | †0.0004 | 8 T, 2 T* 1 L* | 0.0009 | 24 T* |
| Gas Furnace | BP | 0.0159 | 0.0358 | 8 T | 0.0766 | 18 T* |
| | SCG | 0.0110 | †0.0210 | 8 T, 2 T* | 0.0330 | 16 T* |
| | QNA | 0.0115 | 0.0256 | 7 T, 2 L* | 0.0376 | 18 T* |
| | LM | 0.0120 | 0.0223 | 6 T, 1 L, 1 T* | 0.0451 | 14 T* |
| Waste Water | BP | 0.0441 | 0.0547 | 6 T, 5 T*,1 L | 0.1360 | 16 T* |
| | SCG | 0.0457 | 0.0579 | 6 T, 4 L* | 0.0820 | 14 T* |
| | QNA | 0.0673 | 0.0823 | 5 T, 5 TS | 0.1276 | 14 T* |
| | LM | 0.0425 | †0.0521 | 8 T, 1 LS | 0.0951 | 14 T* |

**Table 11.** Performance results and run time comparison of MLEANN

| Time series | Learn Algo. | EANN RMSE Training | EANN RMSE Test | Architecture | Run time in minutes ++A | Run time in minutes +B |
|---|---|---|---|---|---|---|
| Mackey Glass | BP | 0.0166 | 0.0168 | 4T | 1181 | 288 |
| | SCG | 0.0062 | 0.0067 | 3 T, 1 T* | 2066 | 504 |
| | QNA | 0.0059 | 0.0058 | 3 T*, 1 L | 2169 | 528 |
| | LM | 0.0056 | †0.0061 | 2 L*, 2 T* | 2463 | 602 |
| Gas Furnace | BP | 0.0189 | 0.0371 | 3 L | 305 | 62 |
| | SCG | 0.0179 | 0.0295 | 1 T*, 2 L | 629 | 121 |
| | QNA | 0.0156 | 0.0295 | 2 T*, 1 L*, 1 L | 661 | 128 |
| | LM | 0.0181 | †0.0290 | 1 T, 1 L, 1 T* | 696 | 132 |
| Waste Water | BP | 0.0647 | 0.0639 | 2T, 2T* | 702 | 146 |
| | SCG | 0.0580 | 0.0600 | 2 T*, 1 T, 1 L | 1254 | 267 |
| | QNA | 0.0590 | 0.0596 | 3 T*, 1L* | 1291 | 279 |
| | LM | 0.0567 | †0.0591 | 2 L, 1 T, 1 T* | 1176 | 294 |

++ without architecture restriction, + with architecture restriction

† Lowest RMSE error

### 3.3.3 Comparison with Neuro-Fuzzy Systems

In this section we compare the performance of MLEANN (RMSE values on training and test sets) with two popular neuro-fuzzy models. The neuro-fuzzy models [1] considered were Dynamic Evolving Fuzzy neural networks (dmEFuNN) [24] [47] implementing a Mamdani fuzzy inference system [55] and an Adaptive Neuro-Fuzzy Inference System (ANFIS) [42] implementing a Takagi-Sugeno fuzzy inference system [76]. The same training and test sets of the three time series were used to compare the performance with the neuro-fuzzy systems. The empirical results are depicted in Table 12.

**Table 12.** Performance comparison between MLEANN and Neuro-Fuzzy Systems

| Time series | RMSE | | | | | |
| --- | --- | --- | --- | --- | --- | --- |
| | EANN | | Mamdani – NF | | Takagi Sugeno - NF | |
| | Training | Test | Training | Test | Training | Test |
| Mackey Glass | 0.0004 | 0.0004 | 0.0023 | 0.0042 | 0.0019 | 0.0018 |
| Gas Furnace | 0.0110 | 0.0210 | 0.0140 | 0.0490 | 0.0137 | 0.0570 |
| Waste Water | 0.0425 | 0.0521 | 0.0019 | 0.0750 | 0.0530 | 0.0810 |

## 4. Discussions and Conclusions

Table 10 shows comparative performance between MLEANN and a conventional ANN without any architecture restriction. For Mackey glass series, using 500 epochs of BP learning, RMSE on test set was reduced by 82% (BP), 31% (SCG), 29% (QNA) and 56% (LM). At the same time, number of hidden neurons got reduced by approximately 58% (BP), 54% (SCG), 58% (QNA) and 55% for LM. LM algorithm gave the best RMSE error on test set (0.0004) even though it is highly computational expensive as demonstrated in Table 7.

For the gas furnace time series, RMSE on test set was reduced by 53%% (BP), 36% (SCG), 69% (QNA) and 73% (LM). Savings in hidden neurons amounted to 55% (BP), 37% (SCG), 50% (QNA) and 55% (LM). SCG training gave the best RMSE value (0.0210) for gas furnace series.

For the wastewater time series, RMSE on test set was reduced by 60% (BP), 29% (SCG), 35% (QNA) and 45% (LM). Savings in hidden neurons amounted to 25% (BP), 29% (SCG), 29% (QNA) and 36% (LM). LM learning gave the best RMSE value (0.0521) for wastewater series.

To have an empirical comparison, we deliberately terminated the local search after 500 epochs (regardless of early stopping in some cases) for all the training algorithms. In some cases the generalization performance could have been further improved. As depicted in Table 3.4, our experimentations with limited architecture also reveal the efficiency of MLEANN technique. The gas furnace time series and wastewater series could be learned just with 4 hidden neurons using LM algorithm. However, for Mackey glass series the results were not that encouraging when compared with the conventional design using 24 hidden neurons. Perhaps Mackey Glass series requires more hidden neurons to learn the problem within the required accuracy. Table 12 depicts empirical comparison between two popular neuro-fuzzy systems. As evident, MLEANN has outperformed both neuro fuzzy models in terms of the lowest RMSE vales on test set for all the three time series. Selection of the architecture (number of layers, hidden neurons, activation functions and connection weights) of a network and correct learning algorithm is a tedious task for designing an optimal artificial neural network. Moreover, for critical applications and hardware implementations optimal design often becomes a necessity. In this paper, we have formulated and explored; MLEANN: an adaptive computational framework based on evolutionary computation for automatic design of optimal artificial neural networks. Empirical results are promising and show the importance and efficacy of the technique.

In MLEANN, our work was mostly concentrated on the evolutionary search of optimal learning algorithms for feedforward neural networks. Similar approach could be used for optimizing recurrent neural networks and other connectionist networks. For the evolutionary search of architectures, it will be interesting to model as co-evolving [27] sub-networks instead of evolving the whole network. Further, it will be worthwhile to explore the whole population information of the final generation for deciding the best solution. We used a fixed chromosome structure (direct encoding technique) to represent the connection weights, architecture, learning algorithms and its parameters. As size of the network increases, the chromosome size grows. Moreover, implementation of crossover is often difficult due to production of non-functional offspring's. Parameterized encoding overcomes the problems with direct encoding but the search of architectures is restricted to layers. In the grammatical encoding rewriting grammar is encoded. So the success will depend on the coding of grammar (rules). Cellular configuration might be helpful to explore the architecture of neural networks more efficiently. Gutierrez et al [39] has shown that their cellular automata technique performed better than direct coding.

## Acknowledgement

The author wishes to thank the anonymous reviewers for their valuable comments.

## References

[1] Abraham A (2001), Neuro-Fuzzy Systems: State-of-the-Art Modeling Techniques, Connectionist Models of Neurons, Learning Processes, and Artificial Intelligence, Springer-Verlag Germany, Jose Mira and Alberto Prieto (Eds.), Granada, Spain, pp. 269-276.

[2] Abraham A (2002), Optimization of Evolutionary Neural Networks Using Hybrid Learning Algorithms, IEEE 2002 Joint International Conference on Neural networks, IEEE Press, Volume 3, pp. 2797-2802.

[3] Abraham A and Nath B (1999), Failure Prediction Of Critical Electronic Systems in Power Plants Using Artificial Neural Networks", In Proceedings of First International Power & Energy Conference, Isreb M (Editor), ISBN 0732 620 945, Australia, December 1999.

[4] Abraham A and Nath B (2000), Optimal Design of Neural Nets Using Hybrid Algorithms, In proceedings of $6^{th}$ Pacific Rim International Conference on Artificial Intelligence (PRICAI 2000), pp. 510-520.

[5] Abraham and Nath B (2000), Artificial Neural Networks for Intelligent Real Time Power Quality Monitoring Systems", In Proceedings of First International Power & Energy Conference, Isreb M (Editor), ISBN 0732 620 945, Australia, December 1999.

[6] Abraham and Nath B (2001), ALEC -An Adaptive Learning Framework for Optimizing Artificial Neural Networks", Computational Science, Springer-Verlag Germany, Vassil N Alexandrov et al (Editors), San Francisco, USA, pp. 171-180.

[7] Angeline P J, Saunders G B and Pollack J B (1994), An Evolutionary Algorithm that Evolves Recurrent Neural Networks, IEEE Transactions on Neural Networks, Vol. 5:1, pp. 54-65.

[8] Auer P., Herbster M and Warmuth M (1996), Exponentially Many Local Minima for Single Neurons, D Touretzky et al (Eds.), Advances in Neural Information Processing Systems, MIT Press, Cambridge, MA, Vol 8, pp. 316-322.

[9] Baffles P T and Zelle J M (1992), Growing layers of Perceptrons: Introducing the Exentron Algorithm, Proceedings on the International Joint Conference on Neural Networks, Vol 2, pp. 392-397.

[10] Balakrishnan K and Honawar V (1996), Some Experiments in Evolutionary Synthesis of Robotic Neurocontrollers. Proceedings of World Congress on Neural Networks. pp 1035-1040.

[11] Baxter J (1992), The evolution of learning algorithms for artificial neural networks, Complex systems, IOS press, Amsterdam, pp. 313-326..

[12] Bishop C M (1995), Neural Networks for Pattern Recognition, Oxford Press.

[13] Boers E J W, Borst M V and Sprinkhuizen-Kuyper I G (1995), Artificial Neural Nets and Genetic Algorithms, DW Pearson et al, (Eds.); Springer Verlag, NY, Proceedings of the International Conference in Ales, France, pp. 333-336.

[14] Boers E J W, Borst M V and Sprinkhuizen-Kuyper I G (1995); Evolving artificial neural networks using the Baldwin effect, In D.W. Pearson et al (Eds.), Artificial Neural Nets and Genetic Algorithms, Proceedings of the International Conference in Alès, France, pp. 333-336, Springer-Verlag, New York.

[15] Boers E J W, Kuiper H, Happel B L M, and Sprinkhuizen-Kuyper I G (1993); Designing modular artificial neural networks, In: H.A. Wijshoff (Ed.); Proceedings of Computing Science in The Netherlands, pp. 87-96.

[16] Bourlard H.A and Morgan N. (1994), Connectionist Speech Recognition: A Hybrid Approach, Boston: Kluwer Academic Publishers

[17] Box G E P and Jenkins G M (1970), Time Series Analysis, Forecasting and Control, San Francisco: Holden Day.

[18] Branke J, Kohlmorgen U and Schmeck H (1995), A Distributed Genetic Algorithm Improving the Generalization Behavior of Neural Networks, Proceedings of the European Conference on Machine Learning, N Lavrac et al (eds.), pp. 107-112.

[19] Braun H (1995), On Optimizing Large Neural Networks (Multilayer Perceptrons) by Learning and Evolution, in Proceedings of the Third International Congress on Industrial and Applied Mathematics, ICIAM.

[20] Braun H and Weisbrod J (1993), Evolving Neural Networks for Application Oriented Problems, in D.B Fogel (Ed.), Proceedings of the second Conference on Evolutionary Programming, USA.

[21] Braun H and Zagorski P (1994), ENZO-M - a Hybrid Approach for Optimizing Neural Networks by Evolution and Learning, in Y Davidor et al (Eds.), Proceedings of the third Int. Conference on Parallel Problem Solving from Nature, Israel.

[22] Castillo P A, Merelo J J, Prieto A, Rivas V and Romero G (2000), G-Prop: Global optimization of multilayer perceptrons using GAs, Neurocomputing, 35, pp. 149-163.

[23] Chalmers D J (1990), The Evolution of Learning: An Experiment in Genetic Connectionism", In Touretzky D S et al (Eds), Proceedings of the 1990 Connectionist Models Summer School, Morgan Kaufmann, CA, pp. 81-90.

[24] Cherkassky V (1998), Fuzzy Inference Systems: A Critical Review, Computational Intelligence: Soft Computing and Fuzzy-Neuro Integration with Applications, Kayak O, Zadeh LA et al (Eds.), Springer, pp.177-197.

[25] Chong, E. K.P. and Zak, S.H. (1996). An Introduction to Optimization. John Wiley and Sons, Inc. New York.

[26] Cichocki, A. and Unbehauen, R. (1993), Neural Networks for Optimization and Signal Processing. NY: John Wiley & Sons

[27] Darwen P. J (1996), Co-evolutionary Learning by Automatic Modularization with Speciation" PhD Thesis, University of New South Wales.

[28] Duch W and Korczak J (1999), Optimization and global minimization methods suitable for neural networks, Neural Computing Surveys.

[29] Fahlman S E and Lebiere C (1990), The Cascade – Correlation Learning architecture, Advances in Neural Information Processing Systems, D.Tourretzky (Ed.), Morgan Kaufmann, pp. 524-532.

[30] Fine T L (1999), Feedforward Neural Network Methodology, Springer Verlag, New York.

[31] Fogel D (1999), Evolutionary Computation: Towards a New Philosophy of Machine Intelligence, $2^{nd}$ Edition, IEEE press.

[32] Fogel D B (1997), The Advantages of Evolutionary Computation, Bio-Computing and Emergent Computation, D. Lundh, B. Olsson, and A. Narayanan (Eds.), Sköve, Sweden, World Scientific Press, Singapore, pp. 1-11.

[33] Fontanari J F and Meir R (1991), Evolving a learning algorithm for the binary perceptron, Network, vol.2, pp. 353-359.

[34] Forti M (1996), A Note on Neural Networks With Multiple Equilibrium Points, IEEE Transactions on Circuits and Systems-I: Fundamental Theory, 43, pp. 487 (5).

[35] Frean M (1990), The upstart algorithm: a method for constructing and training feed forward neural networks, Neural computations Volume 2, pp.198-209.

[36] Fullmer B and Miikkulainen R (1992), Using Marker-Based Genetic Encoding of Neural Networks To Evolve Finite-State Behaviour, FJ Varela and P Bourgine (Eds), Proceedings of the First European Conference on Artificial Life, France), pp.255-262.

[37] Funabiki N, Kitamichi J and Nishikawa S (1998), An evolutionary Neural Network Approach for Module Orientation Problems, IEEE transactions on Systems, Man, And Cybernetics- Part B: Cybernetics, Vol.28, No.6, pp.849-855.

[38] Grau F (1992), Genetic Synthesis of Boolean Neural Networks with a Cell Rewriting Developmental Process, In D Whitely and Schaffer J D., Proceedings of the International Workshop on Combinations of Genetic Algorithms and Neural Networks, IEEE Computer Society Press, CA, pp. 55.74.

[39] Gutierrez G, Isasi P, Molina J M, Sanchis A and Galvan I M (2001), Evolutionary Cellular Configurations for Designing Feedforward Neural Network Architectures, Connectionist Models of Neurons, Learning Processes, and Artificial Intelligence, Jose Mira et al (Eds), Springer Verlag - Germany, LNCS 2084, pp. 514-521

[40] Harp S A, Samad T and Guha A (1989), "Towards the Genetic Synthesis of Neural Networks", In Schaffer J D (Editor), Proceedings of the Third International Conference on Genetic Algorithms and their Applications, Morgan Kaufmann, CA, pp. 360-369.

[41] Hofgen K U (1993), Computational limits on Training Sigmoidal Neural Networks, Information Processing Letters, Volume 46, pp. 269-274.

[42] Jang J S R (1991), ANFIS: Adaptive Network Based Fuzzy Inference Systems, IEEE Transactions Systems, Man & Cybernetics.

[43] Jayalakshmi G A, Sathiamoorthy S and Rajaram R (2001), An Hybrid Genetic Algorithm – A New Approach to Solve Traveling Salesman Problem, International Journal of Computational Engineering Science, Vol. 2, No. 2, pp. 339-355.

[44] Jones L, The Computational Intractability of Training Sigmoidal Neural Networks, IEEE Transactions on Information Theory, Volume 43, pp 143-173.

[45] Judd S, Neural Network Design and the Complexity of Learning, MIT Press, Cambridge, MA.

[46] Kasabov N (1996), Foundations of Neural Networks, Fuzzy Systems and Knowledge Engineering, The MIT Press.

[47] Kasabov N (1998), Evolving Fuzzy Neural Networks - Algorithms, Applications and Biological Motivation, in Yamakawa T and Matsumoto G (Eds), Methodologies for the Conception, Design and Application of Soft Computing, World Scientific, pp. 271-274.

[48] Kim H B, Jung S H, Kim T G and Park K H (1996), Fast learning method for back-propagation neural network by evolutionary adaptation of learning rates, Neurocomputing, vol. 11, no.1, pp. 101-106.

[49] Kitano H (1990), Designing Neural Networks Using Genetic Algorithms with graph Generation System, Complex Systems, Volume 4, No.4, pp. 461-476.

[50] Kok J N, Marchiori E, Marchiori M and Rossi C (1996), Evolutionary Training of CLP-Constrained Neural Networks, 2nd Int. Conf. on Practical Application of Constraint Technology, pp.129-142.

[51] Liu Y and Yao X (1996), Evolutionary design of artificial neural networks with different node transfer functions, Proceedings of the Third IEEE International Conference on Evolutionary Computation, Nagoya, Japan, pp.670-675.

[52] Liu Y and Yao X (1998), Towards designing neural network ensembles by evolution, Proceedings of the Fifth International Conference on Parallel Problem Solving from Nature (PPSN-V), Lecture Notes in Computer Science, Vol. 1498, AE Eiben, M Schoenauer and HP Schwefel (Ed.), Springer-Verlag, Berlin, pp.623-632.

[53] Mackey MC and Glass L (1977), Oscillation and Chaos in Physiological Control Systems, Science Vol 197, pp.287-289.

[54] Macready W G and Wolpert D H (1997), The No Free Lunch theorems, IEEE Trans. on Evolutionary Computing, vol.1 , no. 1, pp. 67-82.

[55] Mamdani E H and Assilian S (1975), An Experiment in Linguistic Synthesis with a Fuzzy Logic Controller, International Journal of Man-Machine Studies, Vol. 7, No.1, pp. 1-13.

[56] Mascioli F and Martinelli G (1995), A constructive algorithm for binary neural networks: The oil Spot Algorithm, IEEE Transaction on Neural Networks, 6(3), pp 794-797.

[57] Masters, T. (1994), Signal and Image Processing with Neural Networks: A C++ Sourcebook, John Wiley and Sons, Inc., New York.

[58] Merril J W L and Port R F (1991), Fractally Configured Neural Networks, Neural Networks, Vol 4, No.1, pp 53-60.

[59] Mezard M and Nadal J P (1989), Learning in feed forward layered networks: The Tiling algorithm, Journal of Physics A, Vol 22, pp. 2191-2204

[60] Miller G F, Todd P M and Hedge S U (1989), Designing Neural Networks Using Genetic Algorithms, Proceedings of the Third International Conference on Genetic Algorithms, JD Schaffer (Ed), pp. 379-384.

[61] Moller A F (1993), A Scaled Conjugate Gradient Algorithm for Fast Supervised Learning, Neural Networks, Volume (6), pp. 525-533.

[62] Moriarty D E and Miikkulainen R (1997),. Forming Neural Networks through Efficient and Adaptive Coevolution. Evolutionary Computation Volume 5, pp. 373-399.

[63] Omlin C W and Giles C L (1993), Pruning Recurrent neural networks for improved generalization performance, Tech. Report No 93-6, CS Department, Rensselaer Institute, Troy, NY.

[64] Phansalkar V V and Thathachar M A L (1995), Local and Global Optimization Algorithms for Generalized Learning Automata, Neural Computation, 7, pp. 950-973.

[65] Polani D and Miikkulainen R (1999), Fast Reinforcement Learning Through Eugenic Neuro-Evolution. Technical Report AI99-277, Department of Computer Sciences, University of Texas at Austin.

[66] Porto V W, Fogel D B and Fogel L J (1995), Alternative neural Network training methods, IEEE Expert, volume 10, no.4, pp. 16-22.

[67] Refenes, A. (Ed.) (1995). Neural Networks in the Capital Markets. Chichester, John Wiley and Sons, Inc., England.

[68] Schiffmann W, Joost M and Werner R (1993), Comparison of optimized backpropagation algorithms, Proceedings. Of the European Symposium on Artificial Neural Networks, Brussels, M. Verleysen (Ed.), de Facto Press, pp. 97-104.

[69] Sebald A V, Chellapilla K (1998), On Making Problems Evolutionarily Friendly Part 1: Evolving the Most Convenient Representations, The Seventh International Conference on Evolutionary Programming, EP98, San Diego, pp. 271-280.

[70] Sebald A V, Chellapilla K (1998), On Making Problems Evolutionarily Friendly Part 2: Evolving the Most Convenient Representations, The Seventh International Conference on Evolutionary Programming, EP98, San Diego, pp. 281-290.

[71] Sexton R, Dorsey R and Johnson J (1998), Toward Global Optimization of Neural Networks: A Comparison of the Genetic Algorithm and Backpropagation, Decision Support Systems, 22, pp. 171-185.

[72] Sexton R, Dorsey R and Johnson J (1999), Optimization of Neural Networks: A Comparative Analysis of the Genetic Algorithm and Simulated Annealing, European Journal of Operational Research, 114, pp. 589-601.

[73] Shang Y and Wah B (1996), Global Optimization for Neural Network Training, Computer, 29, pp.45-55.

[74] Shukla K K and Raghunath (1999), An Efficient Global Algorithm for Supervised Training of Neural Networks, Computers and Electrical Engineering, 25, pp. 195(2).

[75] Stepniewski S W and Keane A J (1997), Pruning Back-propagation Neural Networks Using Modern Stochastic Optimization Techniques, Neural Computing & Applications, Vol. 5, pp. 76-98.

[76] Sugeno M (1985), Industrial Applications of Fuzzy Control, Elsevier Science Pub Co.

[77] Topchy A P and Lebedko O A (1997), Neural Network training by means of cooperative evolutionary search, Nuclear Instruments & Methods In Physics Research, Section A: accelerators, Spectrometers, Detectors and Associated equipment, Volume 389, no. 1-2, pp. 240-241.

[78] Vidhyasagar M M (1997), The Theory of Learning and Generalization, Springer-Verlag, New York.

[79] Weigend A.S. and Gershenfeld N.A. Eds. (1994) Time Series Prediction: Forecasting the Future and Understanding the Past, Reading, MA: Addison-Wesley.

[80] Whitley D (1995), Modeling Hybrid Genetic Algorithms, Genetic Algorithms in Engineering and Computer Science, G. Winter, J. Periaux, M. Galan and P. Cuesta, Eds. John Wiley, pp: 191-201.

[81] Yao X (1995), Designing Artificial Neural Networks Using Co-Evolution, Proceedings of IEEE Singapore International Conference on Intelligent Control and Instrumentation, pp.149-154.

[82] Yao X (1999), Evolving Artificial Neural Networks, Proceedings of the IEEE, 87(9):1, pp. 423-1447.

[83] Yao X and Liu Y (1997), A new evolutionary system for evolving artificial neural networks, IEEE Transactions on Neural Networks, 8(3), pp. 694-713.

[84] Yao X and Liu Y (1998), Making use of population information in evolutionary artificial neural networks, IEEE Transactions on Systems, Man and Cybernetics, Part B: Cybernetics, 28(3): PP.417-425.

[85] Yao X and Liu Y (1998), Towards designing artificial neural networks by evolution, Applied Mathematics and Computation, 91(1): pp. 83-90.

[86] Zhang X M and Chen Y Q (2000), Ray-guided global optimization method for training neural networks, Neurocomputing, 30, pp. 333-337.